\documentclass[letterpaper, 10 pt, conference]{ieeeconf}
\IEEEoverridecommandlockouts
\usepackage[utf8]{inputenc}
\usepackage[hidelinks]{hyperref}
\usepackage{graphicx}
\usepackage[cmex10]{amsmath}
\usepackage{amssymb}
\usepackage{cite}
\usepackage[caption=false,font=footnotesize]{subfig}
\usepackage[usenames, dvipsnames]{color}
\usepackage{overpic}

\graphicspath{{./figures/}}

\begin{document}
\author{Daniel Biediger$^1$, Luben Popov$^2$, and Aaron T. Becker$^1$}
\title{\LARGE\bf The Pursuit and Evasion of Drones Attacking an Automated Turret
\thanks{This work is in support of ONR-Code 30 OTA \#N00014-18-9-0001. 
This work was supported in part by the National Science Foundation under Grant No.\ \href{http://nsf.gov/awardsearch/showAward?AWD_ID=1553063}{[IIS-1553063]} and
\href{https://nsf.gov/awardsearch/showAward?AWD_ID=1849303}{[IIS-1849303]}.
}
\thanks{$^1$Department of Electrical and Computer Engineering,  University of Houston, Houston, TX 77204 USA        {\tt\small  \{debiediger,atbecker\}@uh.edu}.}
\thanks{$^2$Bloomberg, USA        {\tt\small  lpopov1@bloomberg.net}.}
}
\maketitle
\begin{abstract}
This paper investigates the pursuit-evasion problem of a defensive gun turret and one or more attacking drones. The turret must ``visit" each attacking drone once, as quickly as possible, to defeat the threat. This constitutes a Shortest Hamiltonian Path (SHP) through the drones. The investigation considers situations with increasing fidelity, starting with a 2D kinematic model and progressing to a 3D dynamic model. In 2D we determine the region from which one or more drones can always reach a turret, or the region close enough to it where they can evade the turret. This provides optimal starting angles for $n$ drones around a turret and the maximum starting radius for one and two drones. 
\par
We show that safety regions also exist in 3D and provide a controller so that a drone in this region can evade the pan-tilt turret.
Through simulations we explore the maximum range $n$ drones can start and still have at least one reach the turret, and analyze the effect of turret behavior and the drones' number, starting configuration, and behaviors.

\end{abstract}

\section{Introduction}\label{sec:Intro}

An anti-aircraft automated gun turret versus a group of quadcopters is an asymmetric engagement. 
 While there are eight leading designs for a 
 Close-In Weapons System (CIWS) turret, each costs several million dollars and requires a fixed mount~\cite{Fong2008CIWST}.  In contrast, commercially available quadcopters are easily deployed and each cost less than \$1,000~\cite{CIWZstory}.
Automated turrets are used because the speed of engagements exceeds the reaction speed of humans~\cite{kleinman1974modeling}.
This paper analyzes scenarios where the quadcopters can defeat a turret through numbers and motion strategies.

\begin{figure}[tb]
\center{\includegraphics[width=\columnwidth]
{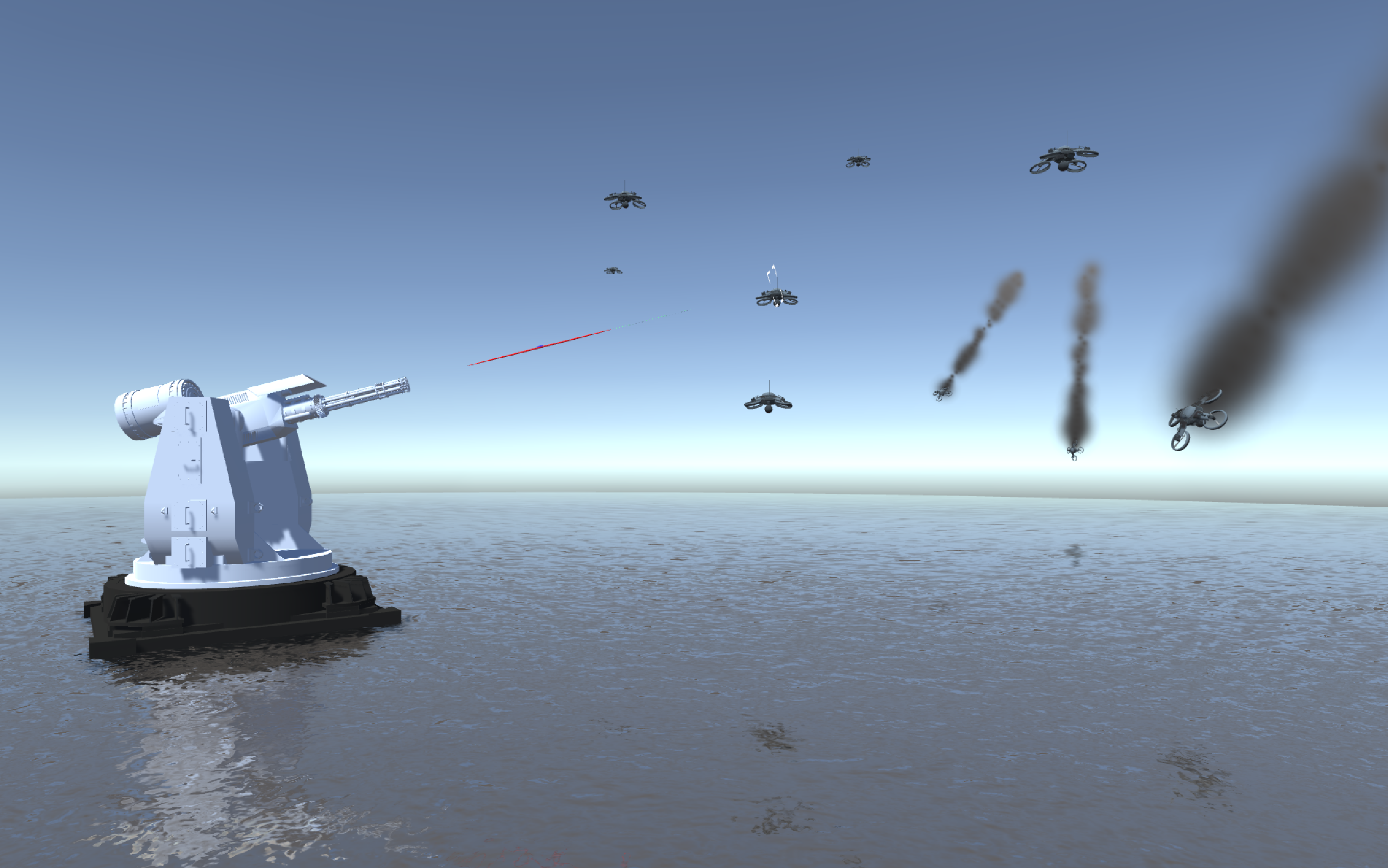}}
\caption{\label{fig:TurretAction} Simulated turret engaging quadcopter attackers.}
\end{figure}

For this paper, a \emph{gun turret} is a stationary defensive weapon with one or more degrees of freedom and kinematic movement (no inertia). 
Initially, we assume that the turret instantaneously kills anything aligned with the turret gun barrel. This models a turret with a rapid rate of fire whose projectiles move much faster than the adversary's top speed, such as a laser turret, or a CIWS. 
Fig.~\ref{fig:TurretAction} shows a frame from our simulation between a turret and seven attacking quadcopters.

This is a pursuit evasion problem where the turret must chase the drones by turning to face them. For a turret with a pan-tilt mount this corresponds to a two-dimensional state space in the pan and tilt angles. The drone evaders have bounded velocity, thrust, inertia and turn rates. To defeat the attack, the turret must visit, in state-space, the pan and tilt angles corresponding to each drone.  

The 2D version of this is similar to the \emph{Lake Monster Problem}, where an evading rowboat in a circular lake must reach the lake shore without being captured by a monster that cannot swim, but can run faster than the rowboat can move~\cite{nahin2012chases}. For both 2D problems, the optimal strategies have a two-step approach that depends on the relative speeds of the agents.

The turret must move through the attacking drones before any drone reaches the turret. To survive, the drones can utilize several strategies. They can attack with overwhelming numbers. They can move into regions where the drones can out-pace the turret motion. They can also split and some move evasively, forcing the turret to follow as long a path as possible, and enabling other drones to reach the turret. Wiener proposed that random evasions are optimal for avoiding anti-aircraft, which was recently confirmed~\cite{kreinovieh2015wiener}. To avoid counter-fire, missiles are  programmed with a variety of strategies~\cite{shinar1995mixed}, including weaving trajectories~\cite{vermeulen2014interception}. This paper compares the effectiveness of the three approaches in 2D and in 3D.

\section{2D Turret \& Drone Theory}

\begin{figure}[tb]
\centering
 \begin{overpic}[width = 0.9\columnwidth]{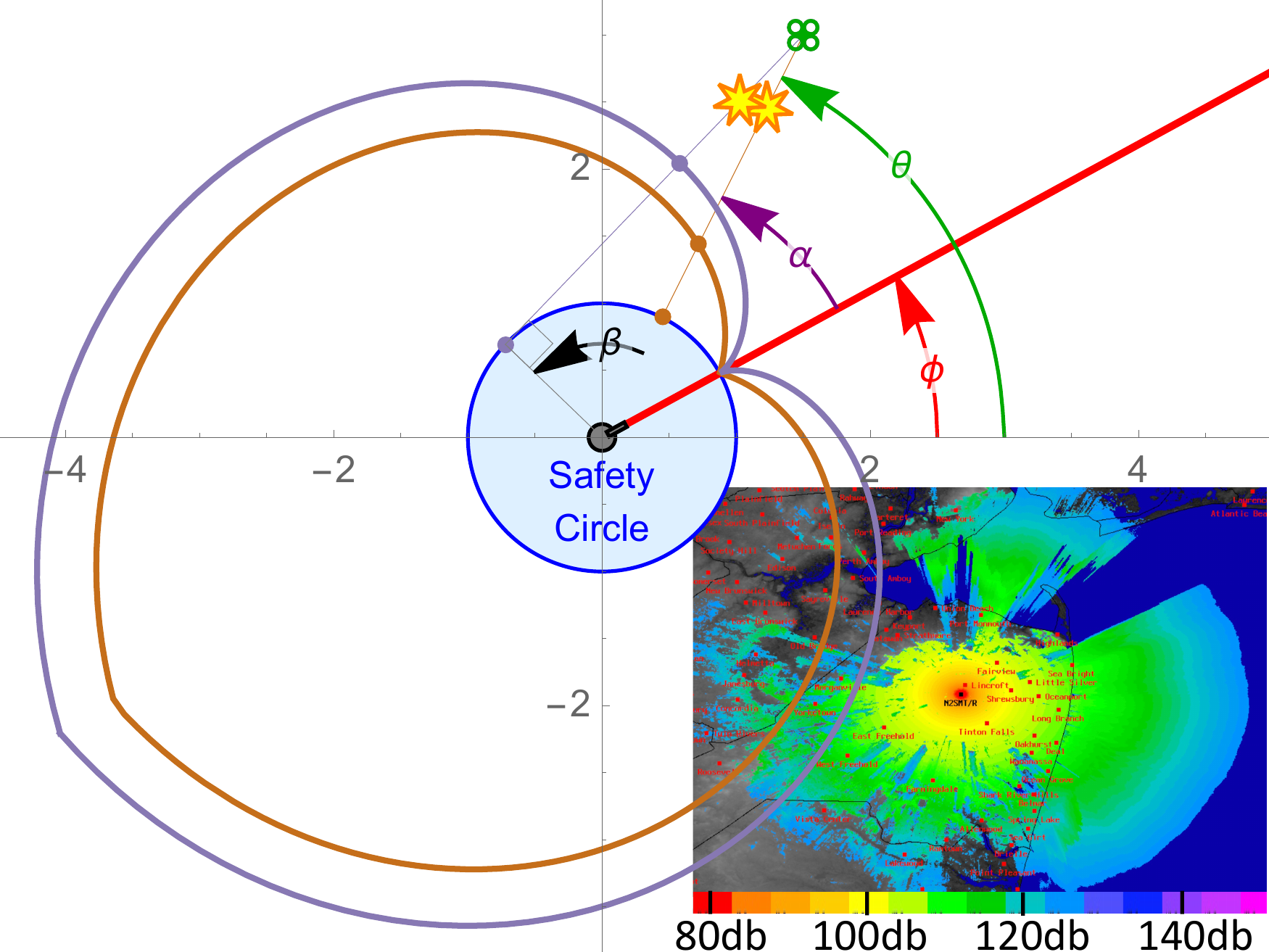}
 \scriptsize
 \put(66,72){$p_0$}
 \put(53,68){$p_d$}
  \put(35,48){$p_\perp$}
\put(42,42){$v$}
\end{overpic}
\caption{\label{fig:Turret2DregionAnnote} 
Schematic of 2D turret, drone, and safety regions. The inset image shows how a drone can exploit terrain masking to avoid detection and stealthily approach a target.}
\end{figure}
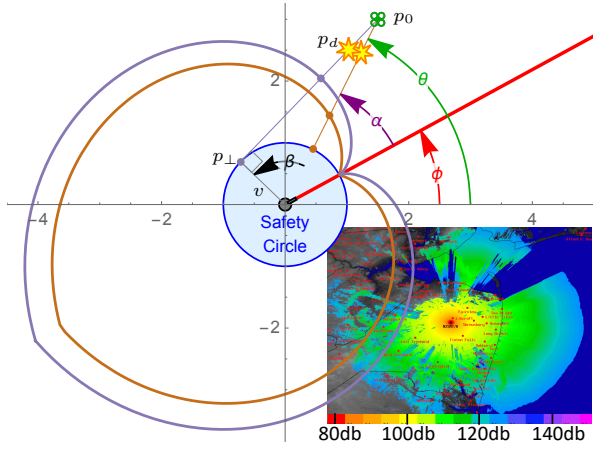

Drone-mitigation methods use
radars and cameras to detect drones.  
To avoid detection, drones can use terrain to mask their approach as long as possible before attempting to reach their target~\cite{RobertsonDroneThreats}. Such an approach can reduce the effective problem from 3D to 2D. 
The effect of terrain masking is shown as an inset to Fig.~\ref{fig:Turret2DregionAnnote}, calculated using the open source SPLAT! software for a 9m radar tower~\cite{SPLAT}. This section identifies the maximum distance from a gun turret where a drone(s) can first be detected and yet still reach the turret.

In a 2D model, the turret's orientation is $\phi$ and the turret has a maximum rate of traverse (turning rate) of $\omega$, $|\dot{\phi}| \le \omega$. We assume that the turret can rotate infinitely in either direction. The mobile opponent is a drone with position  $p\in \mathbb{R}^2$ and has instantaneous velocity  $|\dot{p}| \le v_D$, where $v_D \in \mathbb{R}^+$. To simplify analysis, we eliminate a parameter by dividing the drone velocity by the maximum rate of traverse: $|\dot{p}| \le \frac{v_D}{\omega} = v$.
For the following proofs, we assume that the turret always moves at a top speed of 1 rad/s, so the path length of the turret is equal to the time $t$. For simplicity, we place the turret at $[0,0]$.

A drone located at $[x,y]$ has angle $\theta = \mathrm{arctan}(x,y)$  with respect to the $x$-axis.
The minimum angle from the turret to the drone is $\alpha = \mathrm{arctan}(\cos(\theta-\phi), \sin(\theta-\phi))$.

\subsection{One mobile drone vs.\ a limited rate of traverse turret}

Consider the diagram shown in Fig.~\ref{fig:Turret2DregionAnnote}. The drone initial position is $p_0$ and 
at time $t$ is located at $p_t = [x(t),y(t)]$.
 If the drone is no further than $v$ from the turret, the drone can fly faster than the turret can rotate, since it can easily move 1 rad/s by flying along this circle's circumference.  We call this circle of radius $v$ the ``Safety Circle'', excluding the region where the turret is currently aimed, along the line from $[0,0]$ to $v[\cos(\phi),\sin(\phi)]$.

If the angle from the drone to the turret is $\alpha$, the turret will take $\alpha$ seconds to rotate to the drone's position, in which time the drone can move $v \alpha$ distance.  Therefore, the region bounded by the radius $\rho(\alpha)$ is also survivable:
\begin{align} \label{eq:radial}
   \rho(\alpha) =  v (1 + | \alpha | ), \qquad \ \text{ for } \alpha \in [-\pi,\pi].
\end{align}
This region is outlined by brown in Fig.~\ref{fig:Turret2DregionAnnote}.
If $\|p_t\| > v (|\alpha(t)| +1)$, then the drone cannot reach the safety circle in time by flying radially inwards.  If the drone flies directly toward the origin, it will be intercepted at $(\|p_0\| - v |\alpha|) \left[\cos(\theta), \sin(\theta)\right]$.

However, a larger survivable region, outlined in purple, is possible if the drone flies at an angle instead of flying radially inwards.  The angle that maximizes this region brings the drone to the tangent of the safety circle.
The point tangent to the circle from $p_0$ in the direction of $\alpha$ is
\begin{align}\label{eq:tangentPoint}
   p_\perp &= v \left[\cos (\theta + \beta ),\sin (\theta + \beta)\right], 
   \\
\textrm{where }   \beta &= \textrm{sign}(\alpha) \cos ^{-1}\left(\frac{v}{\|p_0\| }\right).\label{eqn:TangentMeetAngle}
\end{align}
Here $\beta$ is calculated using the 90$^\circ$ triangle $([0,0], p_\perp,p_0)$. 
If the drone starts further than
\begin{align}
    \frac{\left\| p_0-p_\perp\right\| }{v}>\left| \alpha \right| +\left| \beta \right|,
\end{align}
it cannot reach the tangent position, and is destroyed at time $t_d$. This time is computed by numerically solving the following equation:
\begin{align}
    t_d &= \left| \alpha \right| +\left| \beta \right| -\tan ^{-1}\left(\frac{\left\| p_0-p_{\perp}\right\| -t_d v}{v}\right).
    \end{align}
    The drone is destroyed at position
    \begin{align}
    p_d &= p_0- t_d v\frac{ (p_0-p_{\perp})}{\left\| p_0-p_{\perp}\right\| }.
\end{align}

To compute the survivable region when flying to the tangent point of the safety circle, we parameterize the points where the drone ends tangent to the safety circle of radius $v$.
We then solve for the corresponding starting positions:
\begin{align}
   p_0(\gamma) = v \big[&\cos (\gamma +\phi )+\gamma \sin (\gamma +\phi ), \\
   &\sin (\gamma +\phi )-\gamma \cos (\gamma +\phi )\big], \nonumber \\
   \text{for } \gamma \in [-&\gamma_{\max},\gamma_{\max}].\nonumber
\end{align}
The maximum angle occurs when $\tan(\gamma_{\max} ) = \gamma_{\max}$, at $\gamma_{\max} \approx 4.49$ rad.

\begin{figure}[tb]
\center{\includegraphics[width=\columnwidth]
{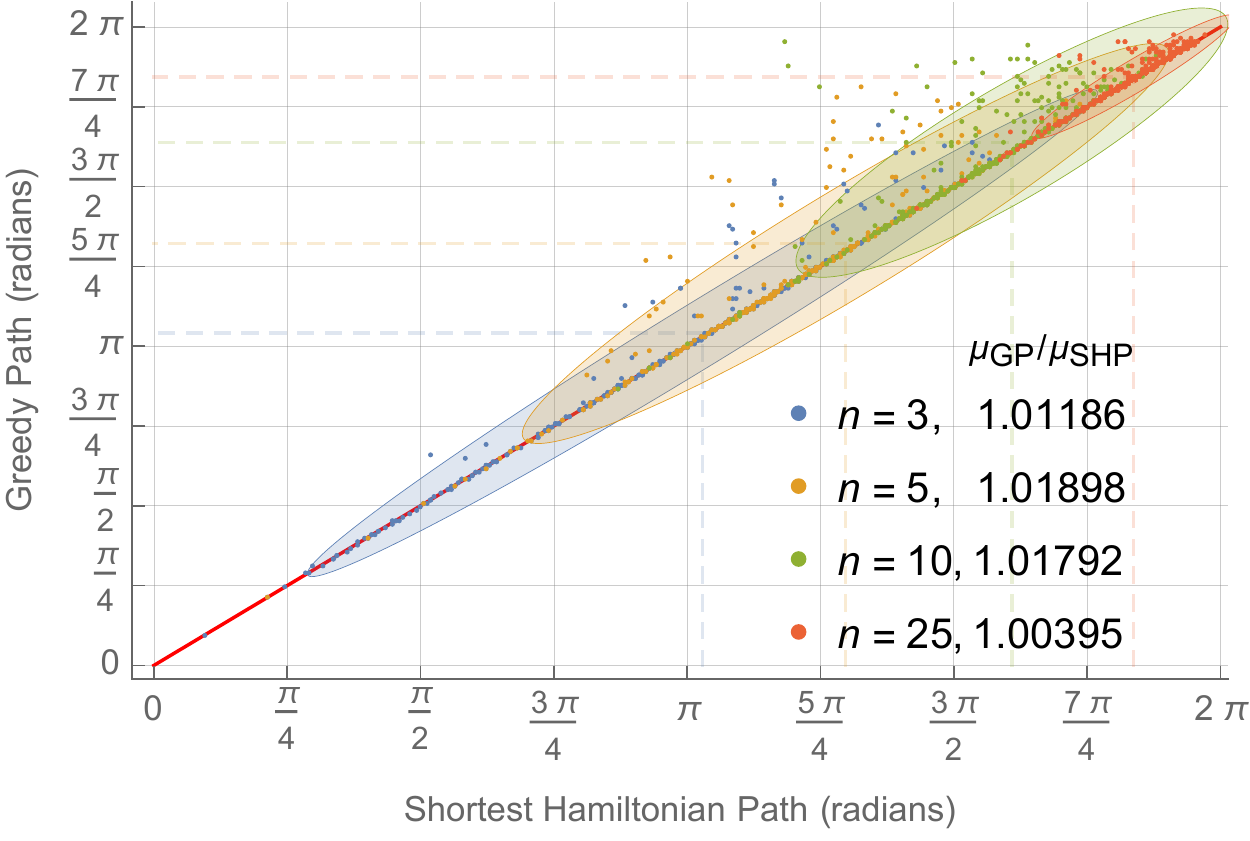}}
\caption{\label{fig:SHPConsordeVsNNin2D} Comparing in 2D the shortest Hamiltonian path and greedy path planning in 2D. 500 trials were run for each number of drones $n$. A 95\% confidence ellipse is plotted for each $n$.}
\end{figure}  

\subsection{Maximizing the turret travel, \texorpdfstring{$n$}{n} radially moving drones}

\begin{figure}[tb]
\centering\hspace{1em}
 \begin{overpic}[width=0.9\columnwidth]{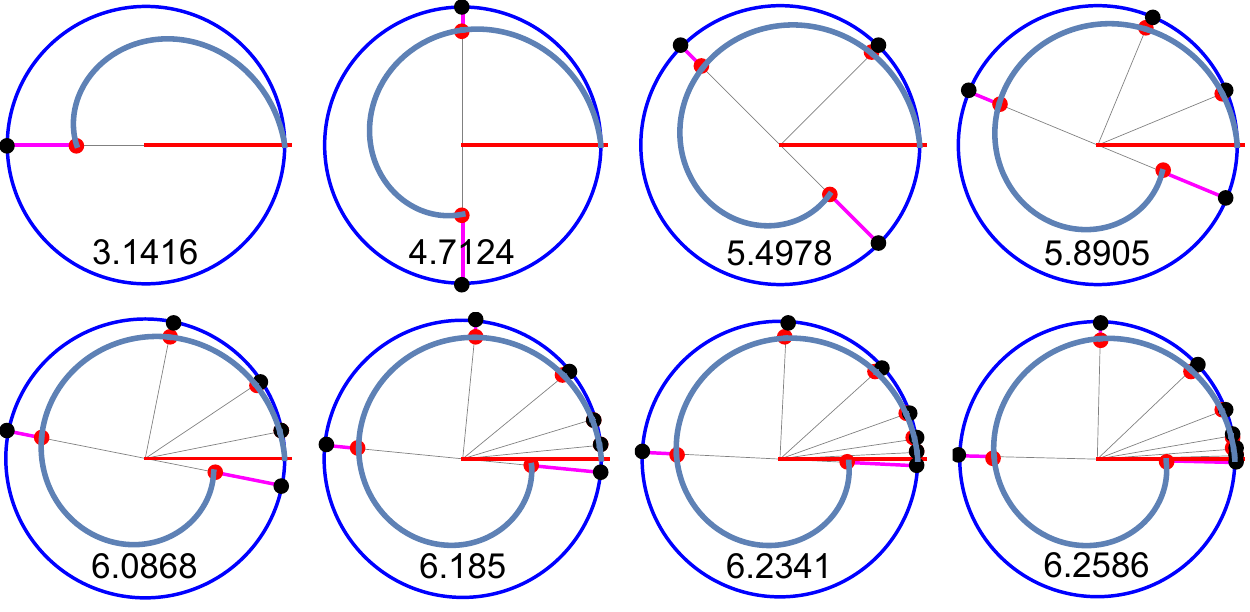}
\put(-5,12){\rotatebox{90}{greedy turret}}
 \end{overpic}\\\centering\hspace{1em}
  \begin{overpic}[width=0.9\columnwidth]{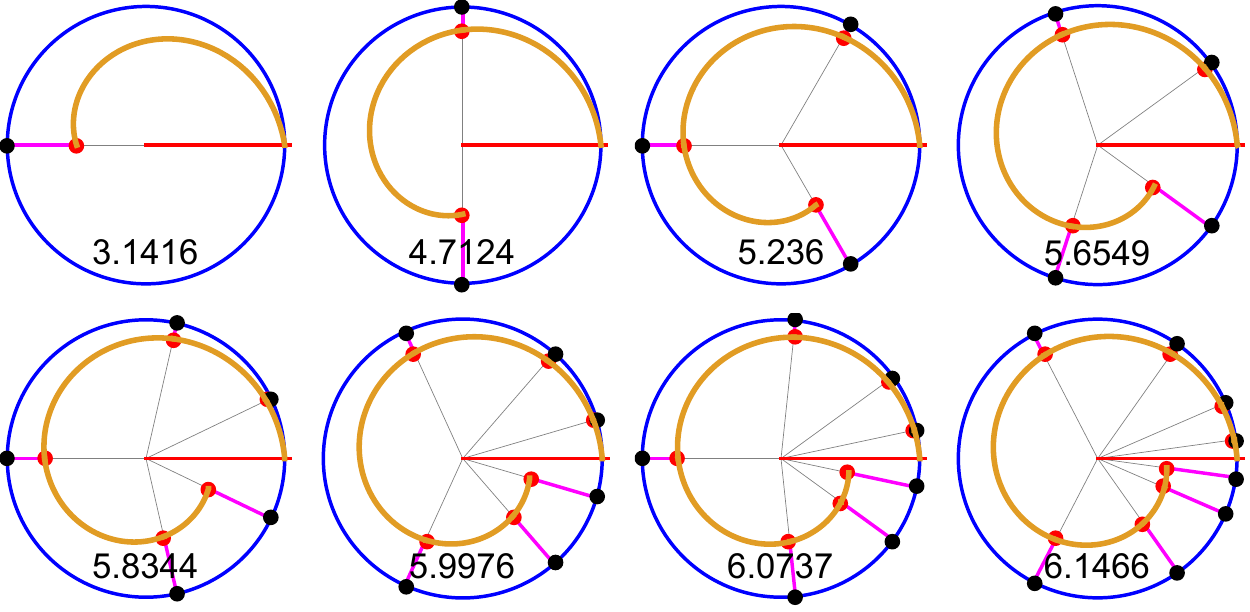}
\put(-5,2){\rotatebox{90}{both, doubling spacing}}
 \end{overpic}\\
  \centering\hspace{1em}
  \begin{overpic}[width=0.9\columnwidth]{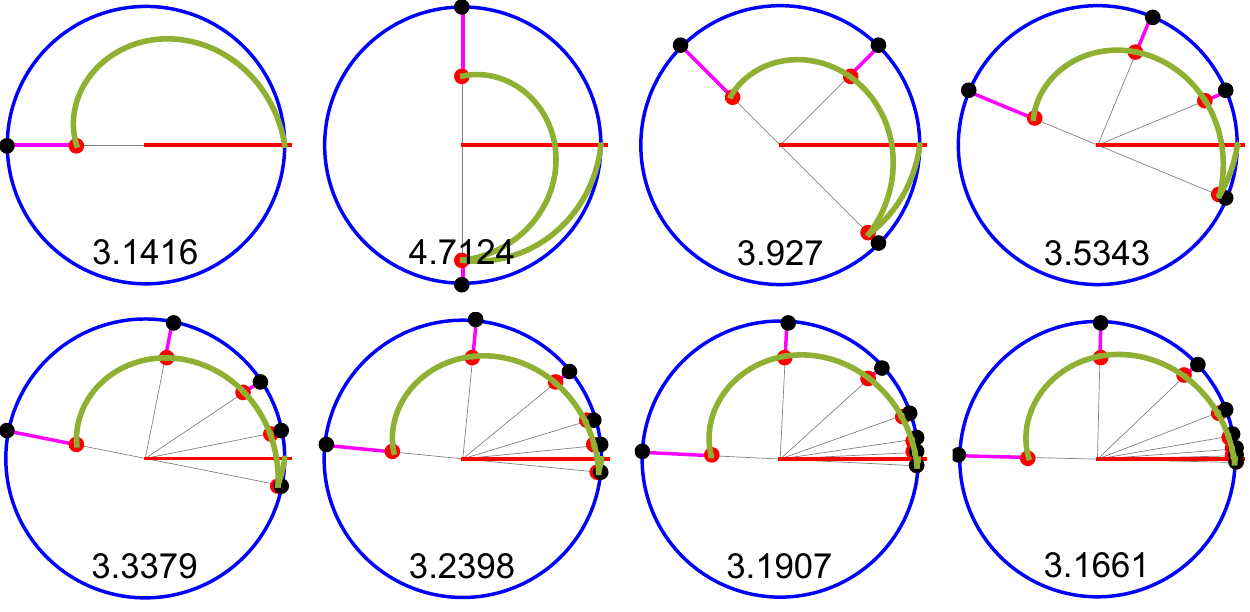}
  \put(-5,12){\rotatebox{90}{optimal turret}}
\end{overpic}
 \begin{overpic}[width=0.9\columnwidth]{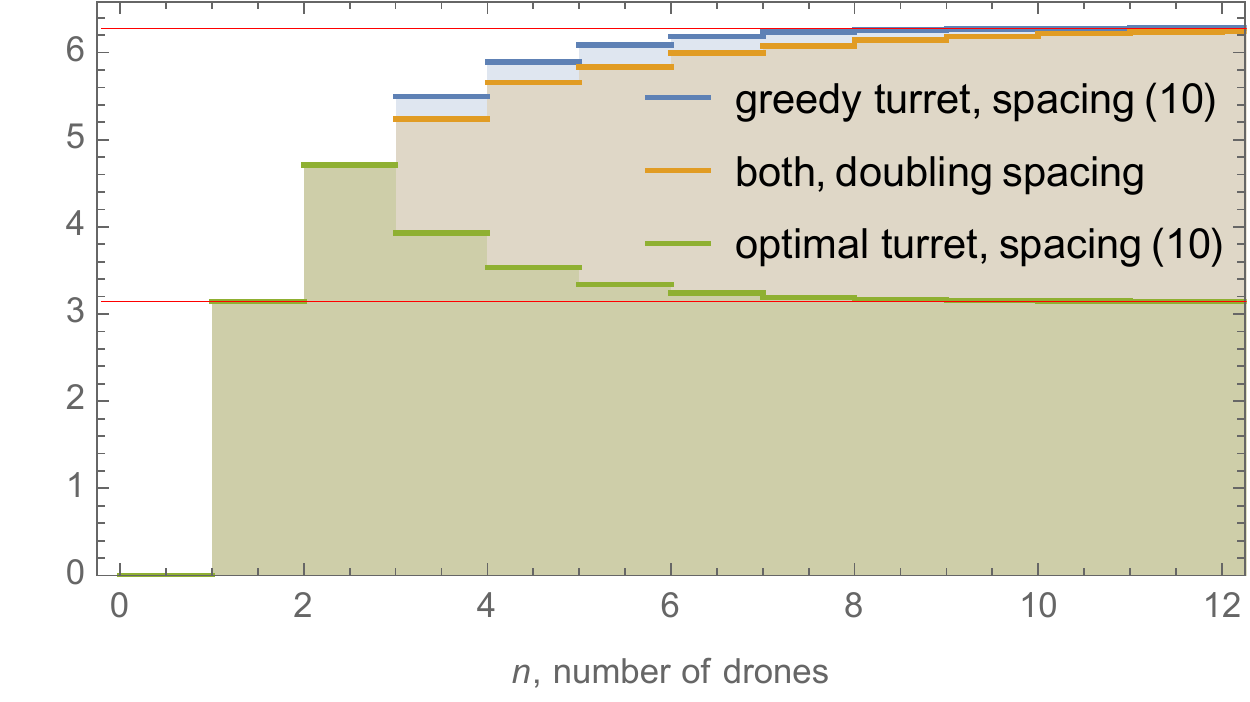}
 \put(-2,8){\rotatebox{90}{turret path length (radians)}}
\end{overpic}
\caption{\label{fig:GreedyTurretRadialPathNdrones} 
Placing drones to maximize path length for a greedy turret results in path lengths that approach 2$\pi$. The same placements for an optimal turret have path lengths that approach $\pi$. Doubling spacing using~\eqref{eq:alphaOptimal} causes both strategies to approach 2$\pi$.
}
\end{figure}
If the drones start too far away to reach the safety circle, they can still work to maximize the time the turret requires to destroy them.
Given $n$ drones that all move radially inwards toward the turret, 
at what angular positions should the $n$ drones be initially placed to force the turret to follow the longest path?  Fig.~\ref{fig:SHPConsordeVsNNin2D} indicates that the longest path is less than $2\pi$ by comparing the resulting path for $n$ drones randomly placed on the circumference of a circle against (1) a greedy turret that always targets the done which requires the smallest traversal, and (2) an optimal turret that selects targets to construct the overall shortest Hamiltonian path.

If the turret is greedy, it can be exploited by placing the $j$th drone at angle
\begin{align}
\alpha_j =    \left(\sum_{i=0}^{j-1} \left(\frac{1}{2}\right)^{n-1-i} \pi \right) - \epsilon \label{eq:spacingToMaxGreedy}
\end{align}
for some small, non-zero constant $\epsilon$.  This
places the first drone at $\left(\frac{1}{2}\right)^{n-1}\pi - \epsilon$ and the last drone at $-\left(\frac{1}{2}\right)^{n-1}\pi - \epsilon$.  Because the counter-clockwise (CCW) drone is $2\epsilon$ closer, the turret initially turns CCW. The greedy turret never turns clockwise (CW) because the intermediate drones are placed at positions that double the distance travelled so far, but are always closer than the CW drone. This
results in a path length that is
\begin{align}
    L_{\textrm{greedy turret, spacing \eqref{eq:spacingToMaxGreedy} }} = 2\pi \left(1-\left(\frac{1}{2}\right)^{n} \right) - \epsilon. \label{eq:BestGreedyPathLength}
\end{align}
  This path performs poorly with a optimal turret, which simply turns CW first, and then CCW, resulting in a path length of 
  \begin{align}
   L_{\textrm{optimal turret, spacing \eqref{eq:spacingToMaxGreedy} }} =  \pi \left(1+\left(\frac{1}{2}\right)^{n-1} \right) - \epsilon.
\end{align}
The turret paths and the total path lengths are shown in Fig.~\ref{fig:GreedyTurretRadialPathNdrones}.

Given an optimal turret, the drones should be placed symmetrically to the left and the right of the turret, with inter-drone angular spacing that doubles for each drone, placing the drones at $\{\pm \alpha_{\textrm{opt}}, \pm 3 \alpha_{\textrm{opt}}, \pm7 \alpha_{\textrm{opt}}, ...\}$.  If $n$ is odd, the final drone should be placed at $\pi$.
\begin{align} \label{eq:alphaOptimal}
\alpha_{\textrm{opt}} = \frac{2 \pi }{2^{\left\lceil \frac{n}{2}\right\rceil }+2^{\left\lfloor \frac{n}{2}\right\rfloor +1}-2}.
\end{align}
The turret is then forced to move only slightly less than \eqref{eq:BestGreedyPathLength}:
  \begin{align}
   L_{\textrm{both, doubling spacing \eqref{eq:spacingToMaxGreedy} }} = 2\pi-\alpha_{\textrm{opt}}.
\end{align}

In all cases, the total total movement of the turret is bounded, no matter the number of drones. Increasing the number of drones is not effective in the 2D case, instead the drones should move in more effective ways.

\subsection{Two mobile drones vs. a limited rate of traverse turret}

\begin{figure}[tb]
\definecolor{hybrid}{RGB}{217, 142, 32}
\definecolor{transition}{RGB}{86, 116, 170}
\definecolor{tangent}{RGB}{147, 181, 70}
\definecolor{mygray}{RGB}{128, 128, 128}
\centering
 \begin{overpic}[width=\columnwidth]{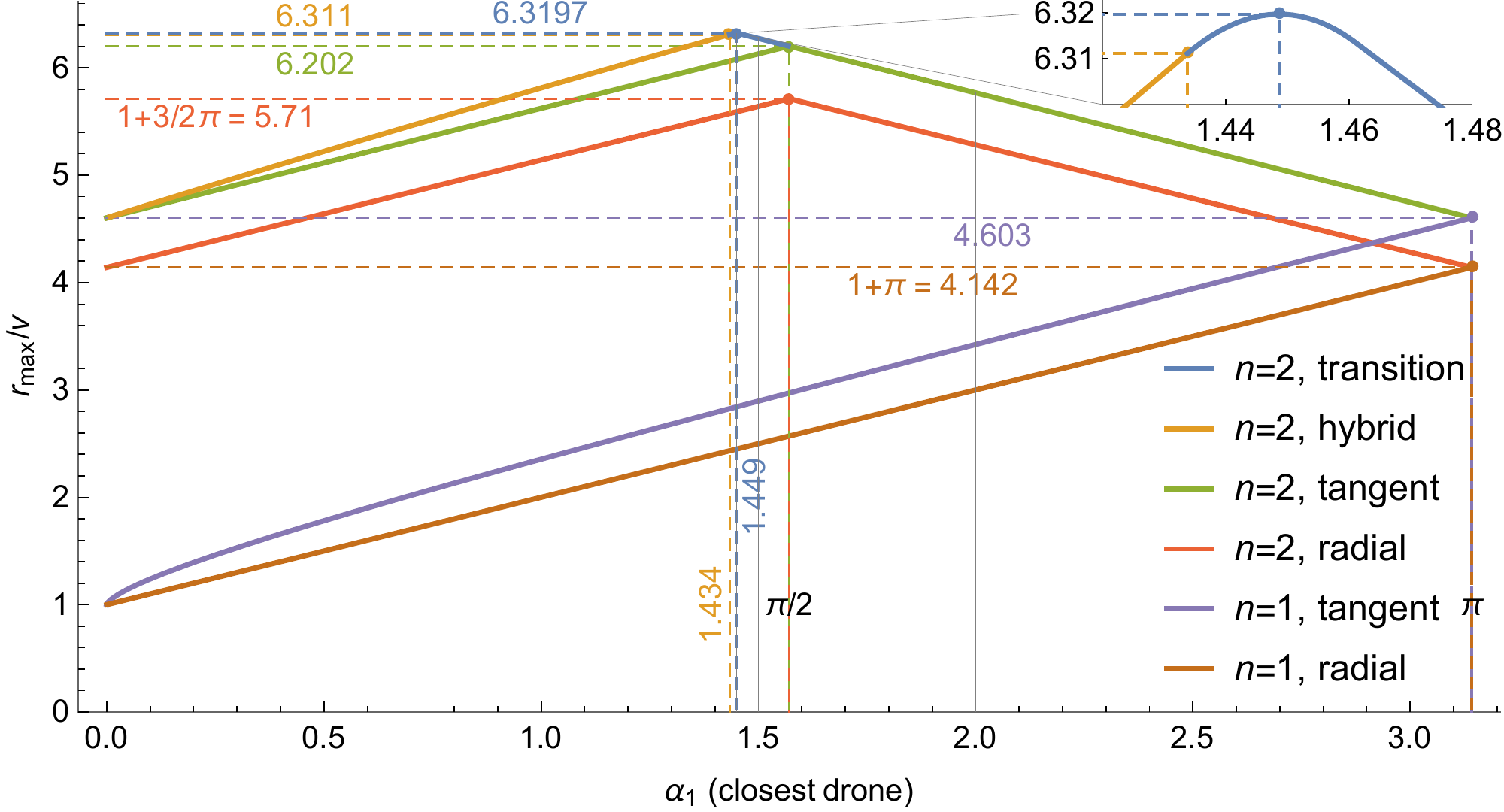}
 \end{overpic}\\
\vspace{1em}
\begin{overpic}[width=0.49\columnwidth]{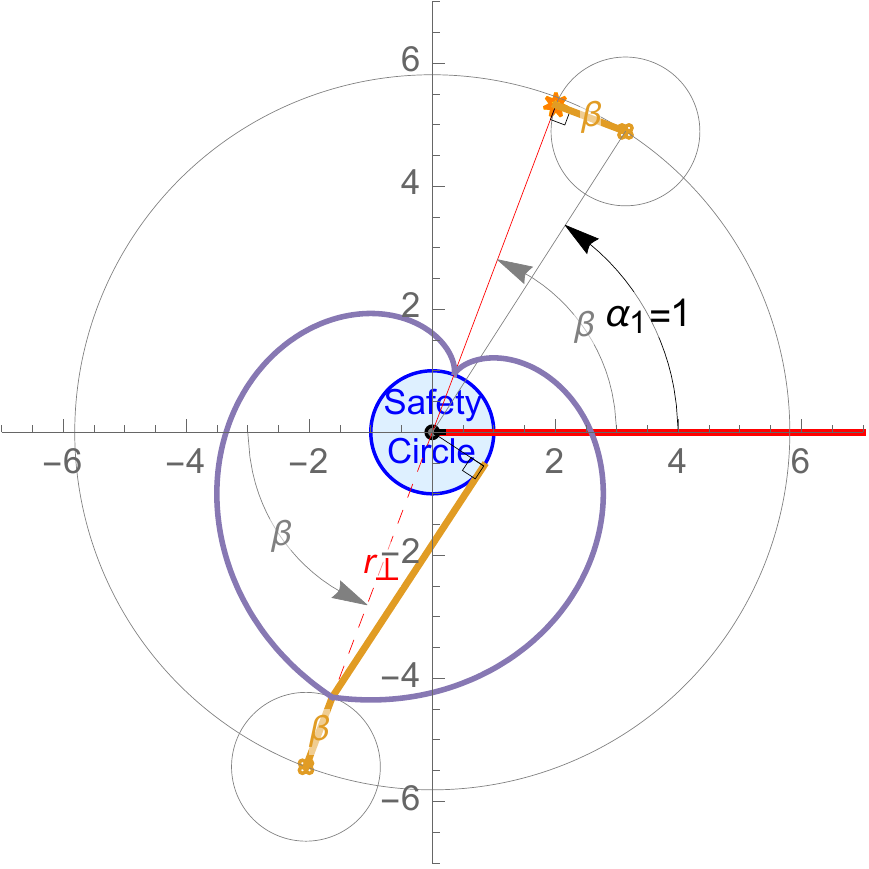}
\put(0,95){\scriptsize A. $\alpha_1 = 1.0$}
\put(0,88){\scriptsize\textcolor{hybrid}{hybrid}}
\put(75,93){\scriptsize\textcolor{mygray}{stay-alive}}
\put(80,88){\scriptsize\textcolor{mygray}{circle}}
\end{overpic}
\begin{overpic}[width=0.49\columnwidth]{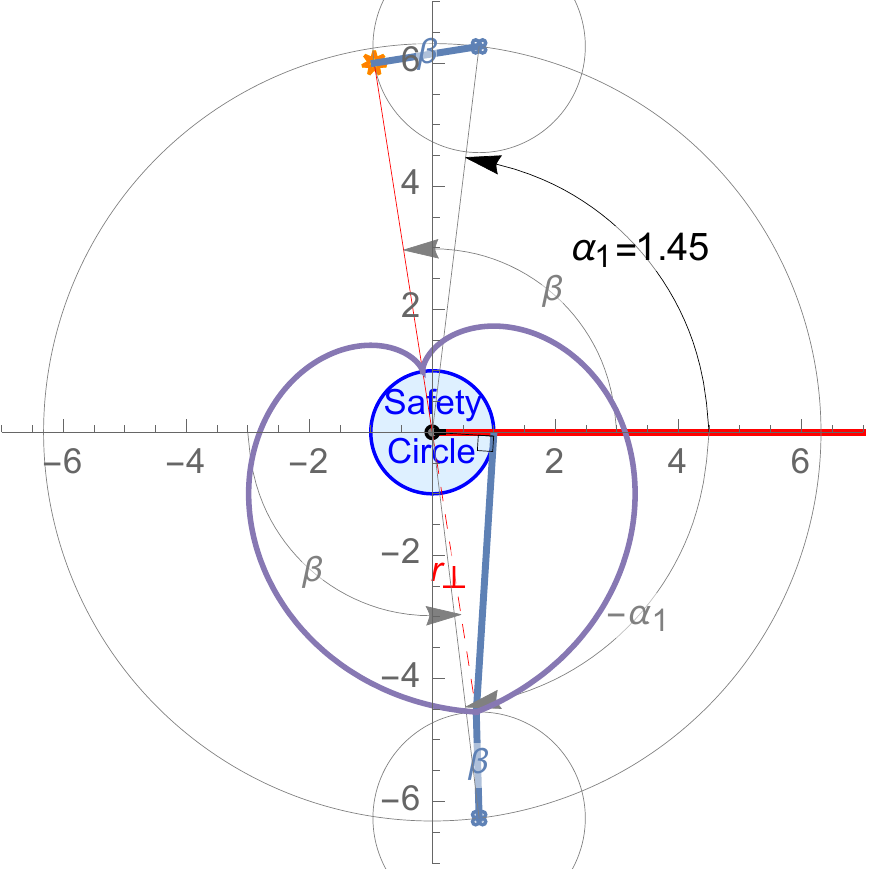} 
\put(0,95){\scriptsize B. $\alpha_1 = 1.45$}
\put(0,88){\scriptsize\textcolor{transition}{transition}}
\end{overpic}
\begin{overpic}[width=0.49\columnwidth]{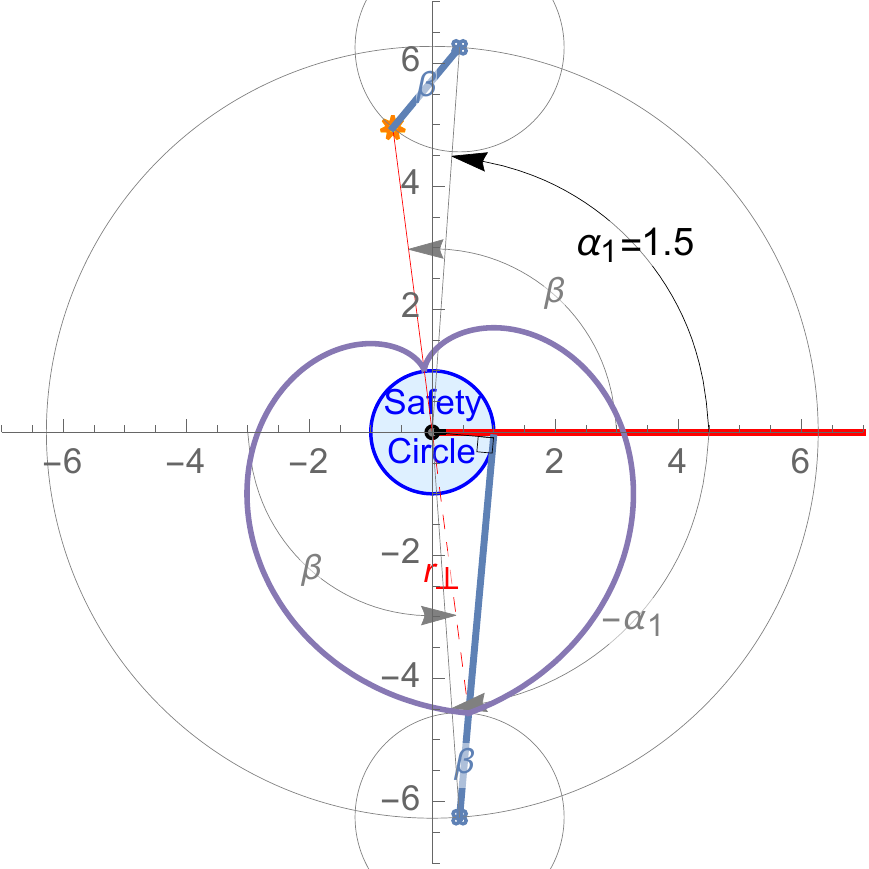}
\put(0,95){\scriptsize C. $\alpha_1 = 1.5$}
\put(0,88){\scriptsize\textcolor{transition}{transition}}
\end{overpic}
\begin{overpic}[width=0.49\columnwidth]{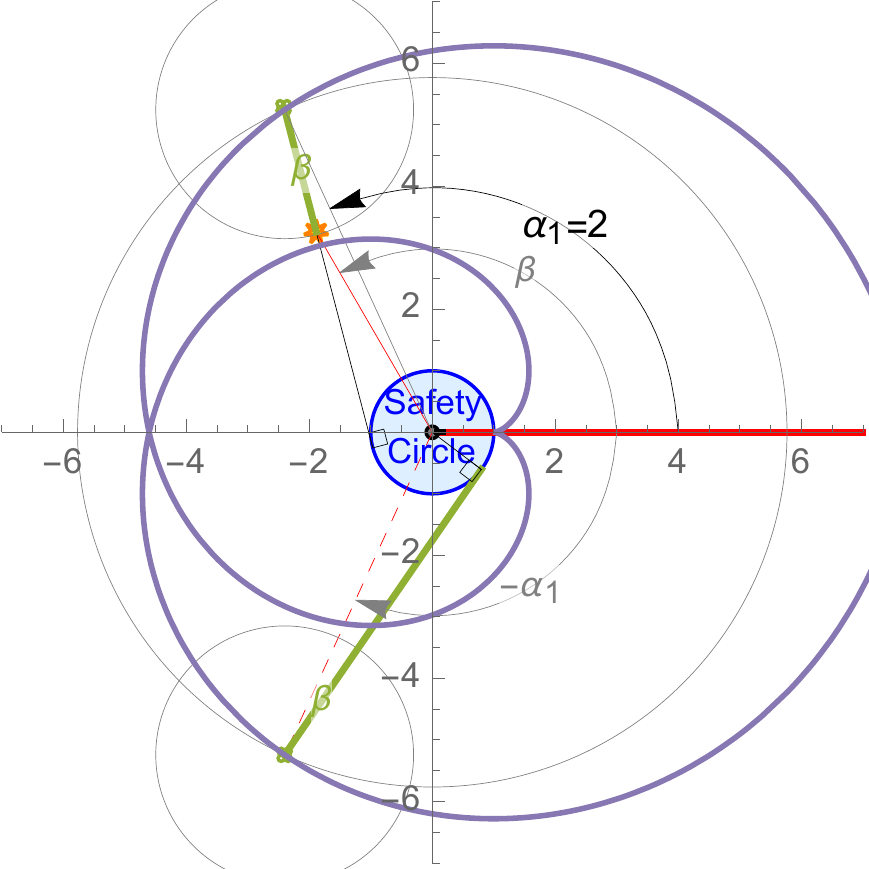}
\put(0,95){\scriptsize D. $\alpha_1 = 2.0$}
\put(0,88){\scriptsize\textcolor{tangent}{tangent}}
\end{overpic}
\caption{\label{fig:OptimalR2drones}(Top) maximum starting distance $r_{\textrm{max}}/v$ with  drone 1 placed at angle $\alpha_1$. (Bottom) optimal paths for $\alpha_1 = \{1,1.45,1.5,2\}$ with 2 drones.}
\end{figure}

Let drone 1 start at radial distance $r$ and starting angle $\alpha_1$ to the turret. We would like to determine the angle $\alpha_2$ and distance $\ge r$ that drone 2 should start, such that drone 2 reaches the safety circle. What motion strategy should each drone follow to maximize $r$? The answer is shown in Fig.~\ref{fig:OptimalR2drones}, which plots $r_\textrm{max}(\alpha_1)$ for several strategies.
First, both drones should start at the same radial distance; starting the second drone further away increases its required time to reach the target, and reduces its effective angular velocity relative to the turret.

For both radially moving drones~\eqref{eq:radial} and for drones that move to the tangent of the safety circle \eqref{eq:tangentPoint}, the second drone should start at $\pi+\alpha_1$ if $\alpha_1 < \pi/2$ (opposite the first drone) and $-\alpha_1$ otherwise. For both, the optimal $\alpha = \pi/2$. 

The maximum radius, $r_\textrm{max}$, for radially moving drones is $(1 + 3\pi/2)v$, and for tangentially moving drones
\begin{align}
r_\perp = r_\textrm{max}(\pi/2) \approx 6.202 v.
\end{align}
For $\alpha_1 \ge \pi/2$, both drones should move toward the tangent of the safety circle, as given by~\eqref{eq:tangentPoint}.

However, for $0\le \alpha_1 \lessapprox 1.434$ there exists a hybrid solution that outperforms the radial and tangent solutions.
Note that the reachable set for drone 1 in time $t$ is a circle of radius $vt$. 
 Drone 1 should move to stay alive as long as possible. This is accomplished by moving in a straight line such that drone 1 dies where the circle of its reachable set is tangent to the firing-line projected from the turret. This ``stay alive circle'' is drawn in light grey in Fig.~\ref{fig:OptimalR2drones}A. The drone will have moved to maximize the angular distance that the turret must travel to pursue it.  
Drone 1's death occurs at angle $\beta$ that solves the equation
\begin{align} \label{eqn:betaToStayAlive}
    \alpha_1 =\beta -\sin ^{-1}\left(\frac{1}{\frac{r_\perp}{\beta }+1}\right).
\end{align}
 Meanwhile, drone 2 starts at angle $\pi +\beta$ and moves radially inward until drone 1 dies. When drone 1 dies, drone 2 is directly behind the turret. The turret can then arbitrarily select to rotate in either direction to engage drone 2. To maximize its survivability,  drone 2 should now move toward the safety-circle tangent point away from the turret's direction of motion.   
 
 At angle $\alpha_1 \approx 1.434$ radians, $\beta + \pi = 2\pi - \alpha_1 $, so for $1.434 \lessapprox  \alpha_1  \le \pi/2$, the second drone cannot start opposite of the place where the drone 1 dies. 
  In this region, the \emph{transition} strategy is optimal.
  Here, drone 1 moves in a straight line but at an angle intermediate between strategy \eqref{eqn:betaToStayAlive} and the tangent meet point. 
  The transition strategy is shown in Fig.~\ref{fig:OptimalR2drones}B and C. The required angle is solved numerically.
 Meanwhile drone 2 starts at $-\alpha_1$ and moves in a straight line to the point $\pi$ radians from  drone 1's eventual death, and then switches to moving toward the tangent to the safety circle, as shown in Fig.~\ref{fig:OptimalR2drones}B. 
 The maximum radius $\approx 6.3197 v$ occurs at $\approx 1.449$ radians, with drone 1 moving at angle $\approx -2.740$ radians with respect to the turret's initial orientation.

\section{3D Turret \& Drone Theory}
\begin{figure}[tb]
\center{\includegraphics[width=\columnwidth]
{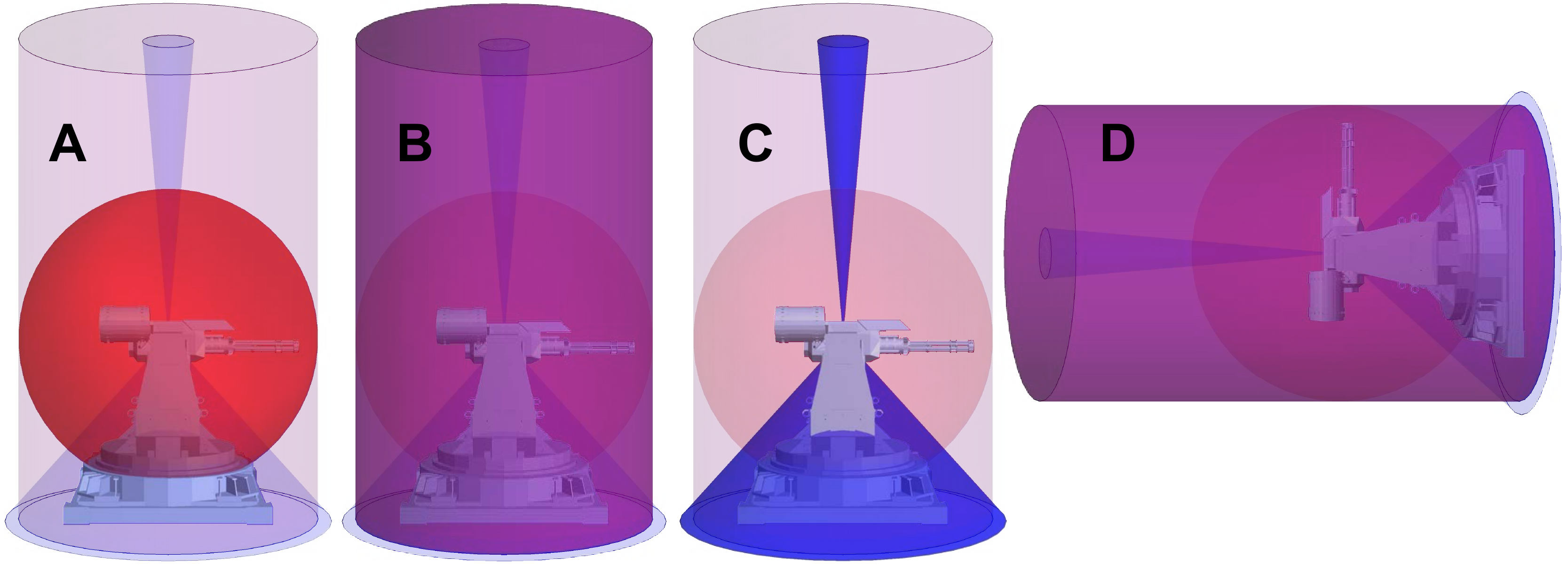}}
\caption{\label{fig:reachableCIWZcomp.pdf} Safety regions for a drone with max velocity 5 m/s and a representative CIWS. The cone and cylinder sets extend infinitely in the axial directions.}
\end{figure}

In 3D there are safety regions that correspond to the 2D safety circle.  
Current CIWS used by five world militaries are degree-of-freedom turrets with a controllable pan and tilt~\cite{digiulianCIWZ,Fong2008CIWST}.
Moreover, the turret has a maximum rate of traverse
$115^\circ$/sec.  For a drone moving with a speed of 5 m/s, this corresponds to a safety circle of radius $\approx$ 2.5 m. 
 For a turret whose speed is constrained by geodesic distance, the safety region is the red sphere in  Fig.~\ref{fig:reachableCIWZcomp.pdf}A, but the turret is actually limited by traverse rate in both pan and tilt, resulting in the cylindrical safety region shown in Fig.~\ref{fig:reachableCIWZcomp.pdf}B. 
 If the tilt angle is limited from $-25^\circ$ to $+85^\circ$, then these rotation limits mean that drones in the blue conical regions shown in Fig.~\ref{fig:reachableCIWZcomp.pdf}C are safe. 
These limits are just mechanical constraints and the ranges vary among CIWS designs~\cite{Fong2008CIWST}, so our analysis will assume no joint limits.
A pan-tilt robot equipped with a gun can be modeled as a spherical robot where the final link is a prismatic joint.  The kinematics are a rotation $R_{z,\theta} R_{y,\phi}$ and a translation $T_d$ along the current $z$-axis. The manipulator Jacobian is then
\begin{align}
J(\theta, \phi,d) =
  \scriptsize  \left[
\begin{array}{ccc}
 -d \sin (\theta) \sin (\phi) & d \cos (\theta) \cos (\phi) & \cos (\theta) \sin (\phi) \\
 d \cos (\theta) \sin (\phi) & d \sin (\theta) \cos (\phi) & \sin (\theta) \sin (\phi) \\
 0 & -d\sin (\phi) & \cos (\phi) \\
\end{array}
\right].\nonumber
\end{align}
This Jacobian has singularities whenever $\phi = \{0,\pi\}$ and also trivially whenever $d$ is zero. A drone can evade the turret indefinitely when near the turret's singularity. This singularity could be moved, but not eliminated by mounting the turret at a different angle, as shown in Fig.~\ref{fig:reachableCIWZcomp.pdf}D.

\subsection{Increasing the number of attacking drones as a strategy}
In 2D increasing the number of drones makes little difference on the effectiveness of the attack, since the distance the turret must travel to clear the area around it is bounded. In 3D, the drones can space out on a manifold. If the drones all fly radially inward to attack a turret, the time required for the turret to defeat the attack depends on the number of drones. For radial motion, the angular distance between the drones does not change from the perspective of the turret. Because a sphere has the largest state space, it represents an upper limit on the path length for a number of drones.

The problem reduces to solving the shortest Hamilton path on a sphere. The SHP is the shortest path that visits all points of interest once from a given starting location. It is related to the Traveling Salesman Problem (TSP), which requires the shortest closed tour through all points of interest. Related problems with TSP on a sphere include \cite{uugur2009genetic}, which used up to 400 points with a genetic algorithm and \cite{eldem2017application} which used up to 400 points with ant colony optimisation. A more complete paper that compares algorithms is~\cite{chen2017hybrid}. 
Making long TSP tours was presented in~\cite{goddyn1990quantizers} based on the theory in~\cite{karloff1989long}.

The distance metric for a freely-rotating turret is the shortest angle between two vectors. Given two drone positions with unit vectors (from the turret) of $\vec{u}$ and $\vec{v}$, the angular distance between them is
\begin{equation} \label{eqn:free-rotation-distace}
    \cos^{-1}\left(\vec{u} \cdot \vec{v} \right)
\end{equation}

\begin{figure}[tb]
\center{\includegraphics[width=0.49\columnwidth]
{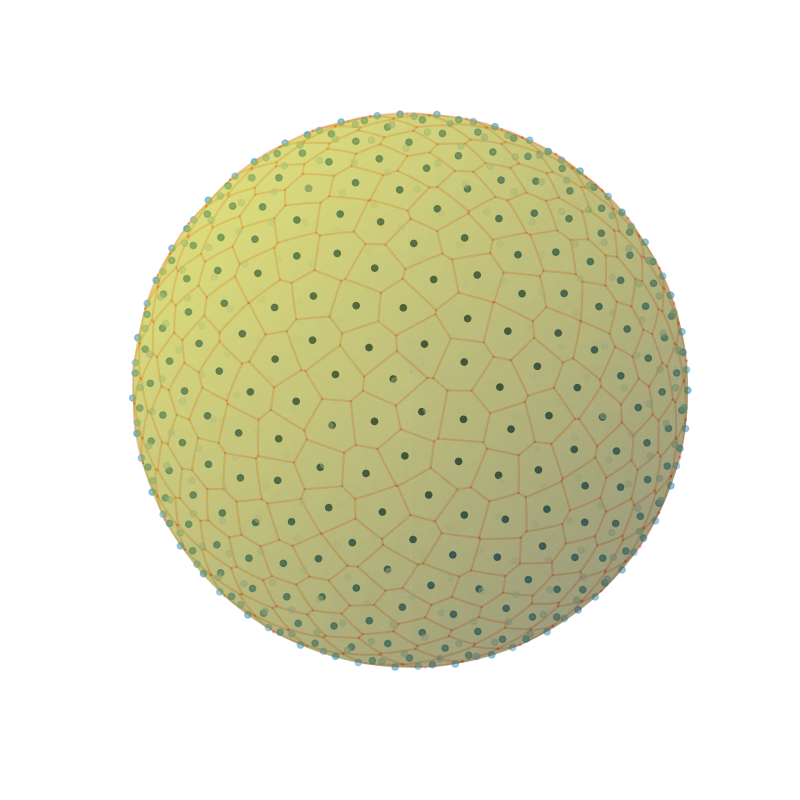}}
\includegraphics[width=0.49\columnwidth]
{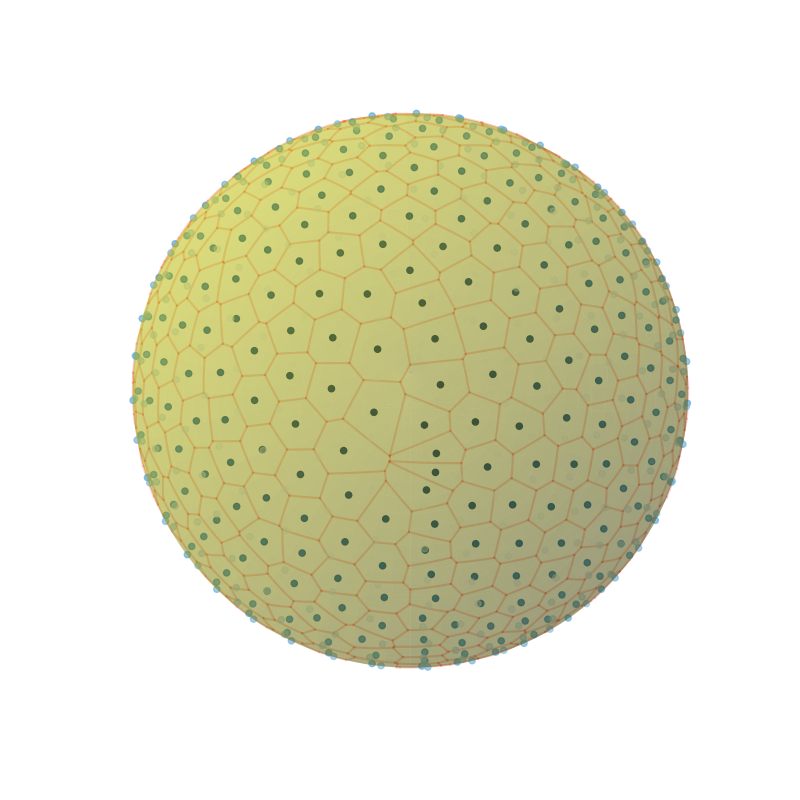}
\vspace{-1em}
\caption{\label{fig:FibonacciSphere} 200 points placed on a sphere using the Fibonacci spiral and randomly generated and spaced with Lloyd's algorithm. }
\end{figure}

To understand the effect of the number of drones on the SHP, we place drones on a sphere using two approaches: a Fibonacci spiral on a sphere~\cite{hardin2016comparison} and random placement with relaxation using Lloyd's algorithm~\cite{lloyd1982least}. The Fibonacci spiral places points on a sphere, using the \emph{golden ratio} to distribute the points somewhat evenly. It produces a spiral pattern on the surface of the sphere, as shown in Fig.~\ref{fig:FibonacciSphere}. Our second approach selects uniformly distributed points on the sphere at random. It then executes Lloyd's algorithm, repeatedly computing the Voronoi cells on the sphere and moving the points to the centroids of these cells. An example of these cells is also shown in Fig.~\ref{fig:FibonacciSphere}. 

\begin{figure}[tb]
\center{\includegraphics[width=\columnwidth]{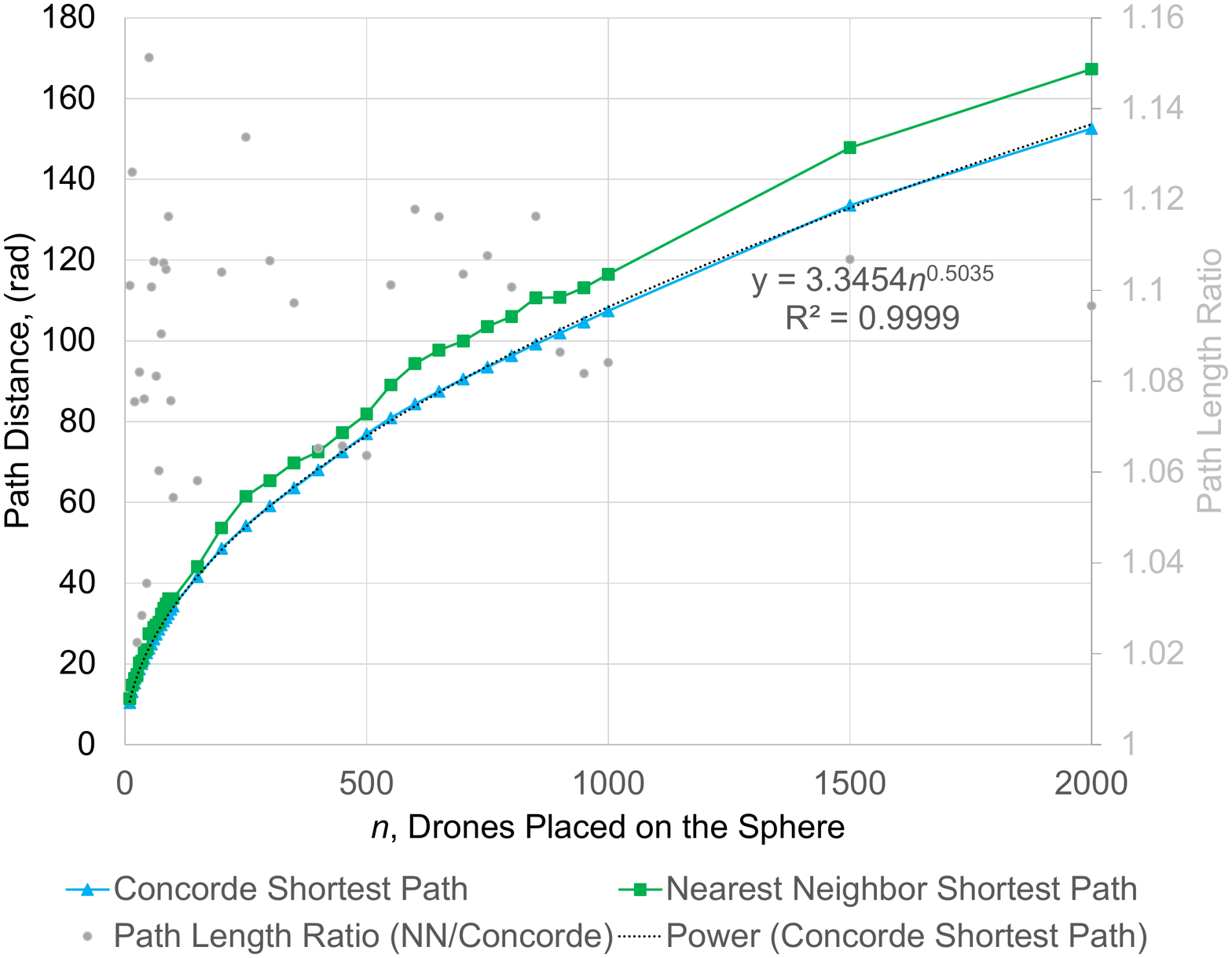}}
\vspace{-1em}
\caption{\label{fig:Compare3DpathSHPvsNN} Nearest neighbor and shortest Hamiltonian path distance comparison for $n$ drones placed using the Fibonacci spiral on a sphere.
}
\vspace{-1em}
\end{figure} 

The length of the path required to engage and defeat all the drones depends on the number of drones present. More drones means more points for the turret to visit in state-space. As the number of points spaced evenly on a sphere increases, the distance between the points decreases. To determine the path length through the points we used two approaches: finding the nearest-neighbor (NN) point at each step and using the freely available \emph{Concord} TSP solver based on the work in \cite{concordePaper}. Concorde has been used to find optimal tours though very large problems including those on the surface of a sphere (with latitude and longitude on Earth) in \cite{applegate2009certification}.\par

We adapted the solver to our problem by adding in an extra phantom point with maximum-distance to and from all points, but zero-distance to and from the starting position. This is inspired by the strategy for transforming asymmetric-TSP problems into symmetric-TSP problems, presented in~\cite{ATSP2TSP}. The solver connects the phantom point to the starting position as either the first or last stop in the tour (the remaining distances are symmetric). To find the SHP, we discard the phantom point and select the direction through the tour with the shortest first move.

The blue line in Fig.~\ref{fig:Compare3DpathSHPvsNN} line shows the SHP computed using the Concord TSP solver. The green line above it shows the results for the finding the path using the NN approach. The point sets used in both cases are generated with the Fibonacci spiral. A best-fit line shows that the path distance increases with the \emph{square-root} of the number of points, $O\left(\sqrt{n}\right)$. This agrees with and extends the results for the distances of a TSP tour on a unit square given by \cite{supowit1983travelling}. Above a small number of drones, the NN approach produces a path that is seven to ten percent longer than the optimal path. For a real turret with more than a few attacking drones, it is not feasible to generate the optimal (and the computationally expensive) solution online. The NN approach provides a fast option for selecting the next target, though it does not find the optimal shortest path above a small number of drones. It also follows a trend where the path length increases  as the $O(\sqrt{n})$ in $n$ drones.

\subsection{Simulating drones attacking a turret}
\begin{figure*}[ht]
         \centering
         \subfloat[][Direct Attack: drones move toward turret]{
             \includegraphics[width=0.49\textwidth]{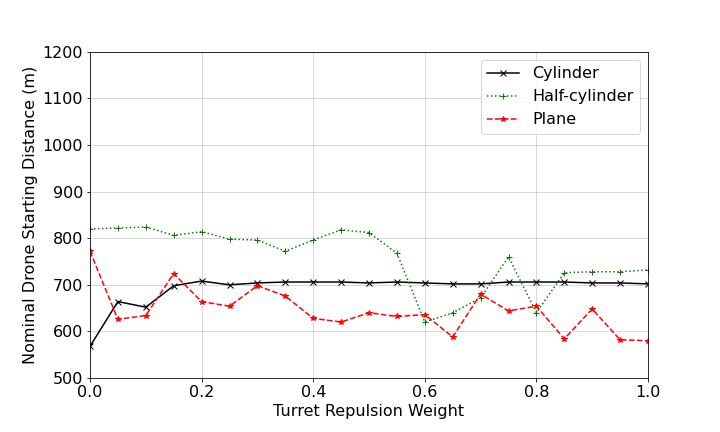}
             \label{fig:DirectAttackComparison}
        }
        \subfloat[][Indirect Attack: drones move toward safety region]{
             \includegraphics[width=0.49\textwidth]{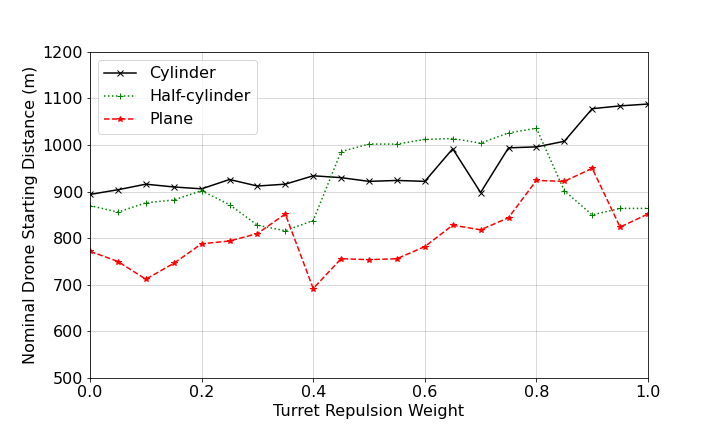}
             \label{fig:IndirectAttackComparison}
        }
\caption{Comparison of drone attack configurations using direct and indirect attack strategies with different levels of turret repulsion.}
\end{figure*} 

\begin{figure}[tb]
\center{\includegraphics[width=\columnwidth]
{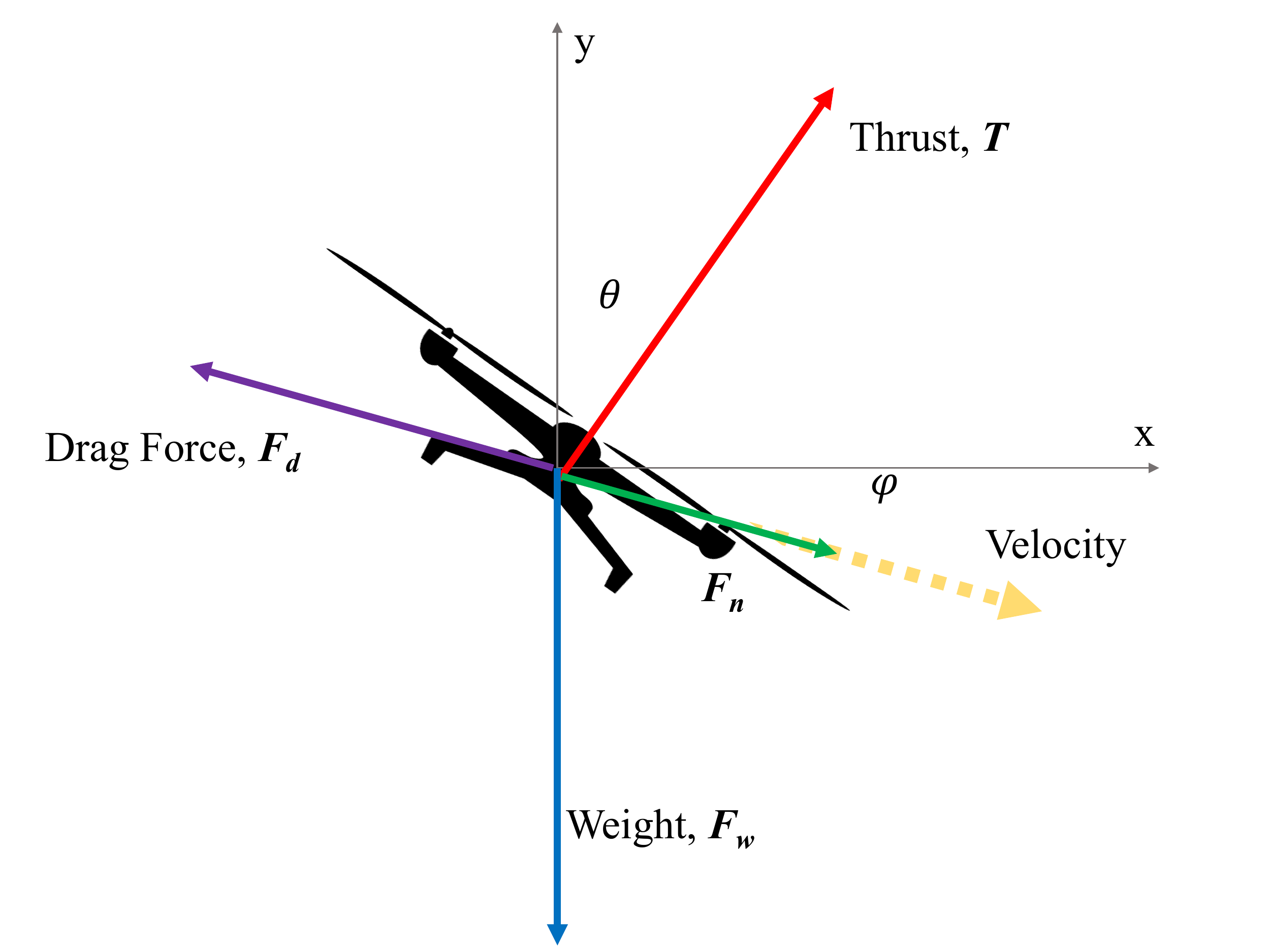}}
\caption{\label{fig:DroneFBD.pdf} Forces acting on a drone.}
\end{figure}

Rather than move radially inward, the drones can attempt to evade the turret by moving with a tangential component relative to the turret. This could potentially increase the traversal distance required by the turret and prolong the life of other drones. In order to test our 3D theories, we used a simulation methodology based on the work in \cite{luben2020Thesis}. The simulation pits the drones against a CIWS turret, with the drones attempting to crash into turret while the turret attempts to eliminate the drones. This framework, built in the Unity game engine, allows us to simulate drones using a dynamic model. This model places limits on the thrust, pitch and yaw rates, top speed, and includes inertia. It can simulate a repulsion between drones to prevent collisions and allows the drones to flee from the turret. Fig.~\ref{fig:DroneFBD.pdf} shows the forces acting on a drone in normal flight. The weight of the drone, $\vec{F_w}$ is counteracted by the thrust $\vec{T}$. Together they sum to a net force $\vec{F_n}$ acting in the desired direction of motion. The figure shows an overlay of the velocity.

\begin{align}
\vspace{-2em}
\vec{F_n} &= \vec{T} + \vec{F_w}
\label{eq:netForce}\\
0 &= \vec{F_n} + \vec{F_d} \nonumber
\end{align}

At maximum speed, the drag force $\vec{F_d}$ counteracts the net force and the drone moves in the direction of $\vec{F_n}$. Without any other repulsive forces, the net force $\vec{F_n}$ is the same as the force pulling the drone to the goal, $\vec{F_g}$.

\subsection{Simulating drones directly attacking a turret with evasion}

Similar to the 2D case with multiple drones, the drones in front of the turret should move in a way that increases the distance the turret must travel. The drones behind the turret should move as quickly as possible toward the turret. We simulate this behavior by placing an outward repulsion from the line extending from the gun on the turret. Drones within a $\pi/4$ cone are subject to the repulsion. 

We place the turret at the origin [0,0,0] and drone $i$ at position $p_i = [x_i, y_i, z_i]$. The unit vector to the drone is $\hat{p_i}$ and the unit vector in the direction that the gun turret it pointing is $\hat{g}$. The direction of drone repulsion is then given by \eqref{eq:gunRepulsionDirection}. The magnitude of this force depends on the angle between the vectors; maximum when the angle between the gun and drone is small and zero when the drone is outside the cone of repulsion.
\begin{align}
\vspace{-2em}
\hat{F_r} &= \left( \hat{g} \times \hat{p_i} \right) \times \hat{p_i}
\label{eq:gunRepulsionDirection}\\
\psi &= \arccos( \hat{g} \cdot  \hat{p_i} ) \nonumber \\
\vec{F_r} &=     \hat{F_r} \left\{
                \begin{array}{ll}
                   1 - \frac{\psi}{\pi/8}  & \psi<\pi/8\\
                  0 & \text{else}
                \end{array}
              \right.
\end{align}

The weight applied to this repulsion, $\xi$, varies in the range [0,1]. A repulsion of zero indicates that the drones ignore the danger posed by the gun and focus only on reaching it. A repulsion of one means that the drones flee the turret and ignore reaching it. In between, the desired net force is a linear combination of the force vector pointing to the goal, $\vec{F_g}$ and the repulsion vector $\vec{F_r}$ given in \eqref{eq:netDesiredForce}.

\begin{align}
\vspace{-2em}
\vec{F_n} = \left( 1 - \xi \right) \vec{F_g} + \xi\vec{F_r}
\label{eq:netDesiredForce}
\end{align}

We generate three drone attack configurations with successively increasing spatial coverage: a plane, a half-cylinder, and a cylinder. The planar formation distributes the $n$ drones in a starting rectangle along staggered rows. The starting region is a distance $d$ in front of the turret and extends from $-\pi/4$ to $\pi/4$ in the turret's pan angle and $0$ to approximately $\pi/4$ in the tilt angle. The half-cylinder places the same $n$ drones evenly spaced in staggered rows from $-\pi/2$ to $\pi/2$ at a distance $d$ in front of the turret. The cylinder surrounds the turret with staggered rows of drones with radius $d$.

To compare the effectiveness of different strategies, we simulate an attack and record its success or failure. If at least one drones reaches (and destroys) the turret, we increase the nominal starting distance and try again. If the drones are defeated, we decrease the starting distance and try again. After successive iterations, we can determine the maximum starting distance for each combination of configurations and strategies. A longer distance reflects an attack that evades and survives for more time. As the maximum distance increases, it is more likely that the drones could approach undetected and complete a successful attack. Shorter distances indicate that the attack is less effective and must start closer to the turret.

Fig.~\ref{fig:DirectAttackComparison} shows a comparison of the drone starting distances for each of the attack patterns. The solid black line shows that the cylinder, beyond some initial benefit from repulsion is relatively unaffected by adding more. The half-cylinder and plane are the most impacted by adding additional repulsion, but the effect is negative. The results are deterministic, but depend on the details of the repulsion and the targets picked by the turret. As the repulsion levels increase, the drones are pushed further away from paths toward the turret. This causes a difference in path the turret takes through the drones.

A high repulsion tends to herd the drones away from the turret making them less effective. The plane suffers the most from the herding effect, as the drones at the sides of the configuration are forced out into unoccupied regions. While this will cause the turret to pan more to reach these drones, they are starting from a much greater radial distance and take longer to reach the turret. The half-cylinder can benefit somewhat from the repulsion to move into unoccupied areas. Low repulsion weights do not negatively impact the performance, but high weights hamper the performance.

\subsection{Simulating drones exploiting the 3D safety region}

If a drone can position itself directly above the turret, or nearly so, it can spiral down avoiding fire from the turret. This corresponds directly to the safety circle in the 2D case where the turret cannot pan fast enough to directly target the drone. We investigated the same three drone attack configurations as before: a plane, a half-cylinder, and a cylinder. To exploit the safety region, the drones move inward toward this cylindrical region above the turret. They move at maximum speed in level flight until they reach the region and then begin to spiral down. The drones attempt to avoid the front of the turret by moving toward a goal behind the gun and below their current position. This position is determined by \eqref{eq:EvadeTurretForever}. In the simulation, $k_1 = 4$m, $k_2 = 5$m and the drone's goal position is $p_{i,\text{goal}}$.
 \begin{align}
     p_{i,\text{goal}} =  [-k_1\hat{g}_x,-k_1\hat{g}_y, p_{i,z} -k_2 ]
     \label{eq:EvadeTurretForever}
 \end{align}

Fig.~\ref{fig:IndirectAttackComparison} shows a comparison of the drone starting distances for each of the attack patterns. The solid black line shows that the cylinder benefits greatly from additional repulsion. In comparison to the direct attack, the indirect attack performs significantly better across all weights. The half-cylinder and plane also benefit from the repulsion and both perform better than under direct attack. 
The plots show more discontinuities in starting distance as well. Because the spiral attack from above is so effective, if drones can reach and enter the safety region they are more likely to defeat the turret. If the turret selects targets in high latitudes, it can counter this threat. In some cases the NN approach does find this path, but in other cases, the turret does not select these drones.  

\section{Conclusions and Future Work}
In 2D, no arrangement of drones can force the turret to follow a path longer than $2\pi$. In a similar way, in 3D increasing the number of drones in an attack to $n$ provides only an $O(\sqrt{n})$ increase in survivability. While avoiding the turret in 2D provides some benefit, the use of repulsion in our 3D simulation provided mixed results. When the drones are in direct attack, the repulsion seems to herd the drones away from the turret more than it increases the survivability. As a result, the overall performance decreases with increasing repulsion. When the drones pursue indirect attack, some turret repulsion increases the survivability of the attack. Repulsion spreads the drones to fill in gaps or to move to empty regions in the airspace. Overall the indirect attack plan performed better than direct attack.

In general, the indirect attack approach performs better than direct attack. If drones can reach the safe region above a turret, they spiral down and destroy it. If the turret tilts up and clears the drones at high latitudes, the attackers near the horizon move in quickly to defeat the turret. To better counter these threats, we plan to improve the turret's target selection to consider not just the nearest target, but also the most proximate threat to the turret or its safety region.  

\bibliographystyle{IEEEtran}
\bibliography{IEEEabrv,turret.bib}
\end{document}